# Applying Deep Bidirectional LSTM and Mixture Density Network for Basketball Trajectory Prediction

Yu Zhao[a], Rennong Yang[a], Guillaume Chevalier[b], Rajiv C. Shah[c], Rob Romijnders[d]

[a] Air Force Engineering University, Xi'an, China

[b]2325 Rue de l'Universite, Laval University, Quebec G1V 0A6, Canada

[c]University of Illinois, Chicago, United State

[d]Eindhoven University of Technology, Eindhoven, Netherlands

**Abstract**

Data analytics helps basketball teams to create tactics. However, manual data collection and analytics are costly and ineffective. Therefore, we applied a deep bidirectional long short-term memory (BLSTM) and mixture density network (MDN) approach. This model is not only capable of predicting a basketball trajectory based on real data, but it also can generate new trajectory samples. It is an excellent application to help coaches and players decide when and where to shoot. Its structure is particularly suitable for dealing with time series problems. BLSTM receives forward and backward information at the same time, while stacking multiple BLSTMs further increases the learning ability of the model. Combined with BLSTMs, MDN is used to generate a multi-modal distribution of outputs. Thus, the proposed model can, in principle, represent arbitrary conditional probability distributions of output variables. We tested our model with two experiments on three-pointer datasets from NBA SportVu data. In the hit-or-miss classification experiment, the proposed model outperformed other models in terms of the convergence speed and accuracy. In the trajectory generation experiment, eight model-generated trajectories at a given time closely matched real trajectories.

## 1 Introduction

Basketball has hundreds of years of history. Today, it is one of the core competitions in the Olympic

Games. There is an increasing trend toward participation in basketball. In professional competitions, each team has a professional coaching team, who design a scientific training plan. Three-point shots, which are always the key to victory, require very important technical tactics. For players, mastering the three-pointer can greatly increase scoring opportunities. In the NBA game, the SportVu player tracking analysis system records everything using sensors and has thus replaced traditional analysis methods. SportVu can collect quantitative statistics regarding player performance, which is useful for post-evaluation and tactical decision-making. Even though the prediction of the basketball's trajectory is a toy problem, this model could contribute to progress in many other areas, for example, behavior recognition [1]-[3], robot control [4][5], or path planning of unmanned aerial vehicles (UAV) [6][7].

The study of a basketball trajectory system can be approached in two ways: i) using mechanical models [8]-[11] or ii) using statistical models [12]-[14]. Mechanical models are simple and easy to use, but they require a solid theoretical foundation and usually require assumptions that limit the scope of application. Statistical models have high generalizability, but require a lot of data and rich feature-processing experience. Moreover, they usually require high computing power and long runtimes. In the 1990s, Hamilton and his collaborators establish a free basketball model based on kinematics that can calculate the best shot angle and speed [8]. Based on previous work, Tran and Silverberg [12] also considered the release height, side angle, and back spin. They concluded that players should aim the ball toward the back of the ring. As long as is the shooter is not covered by an opponent, the shooting height makes no differences to the hit rate. In recent years, people have tried to apply deep learning to trajectory prediction. In a study by Wang and Zemel[13], recurrent neural networks (RNNs) did not perform well in predicting players' motion due to the role of subjective consciousness. However, it achieved good results in classifying players' roles. Subsequently, Zheng [15] used deep hierarchical networks to model player macro-goals and micro-actions. He demonstrated significant improvement over non-hierarchical baselines in the experiment. But his model did not consider competition and cooperation, and it only used an imitation learning framework. Compared with predicting players' motion, the three-dimensional basketball's trajectory is much more complex. Shah and Romijnders [14] proposed an LSTM + MDN model for classifying hit-or-miss outcomes. In contrast to the conventional methods, such as the general linear model (GLM) and the gradient boost model (GBM), LSTM + MDN does not initially require feature extraction. In their paper, GLM and GBM could incorporate only the previous point in generating the trajectory sequence, whereas the proposed model can incorporate the whole sequence, which includes temporal information. The results show that using LSTM+MDN significantly improves hit or miss classification.

BLSTM is a variant of the RNN model, and its structure is particularly suitable for solving time series problems. Unlike LSTM, BLSTM can use forward and backward information. Graves and Schmidhuber [16] firstly proposed BLSTM in 2005, applying it for framewise phoneme classification. Since then, BLSTMs have shown state-of-the-art performance in speech recognition [17] [18], natural language processing [19][20] and other areas [21][22]. In fact, researchers often use deep BLSTMs due to their strong learning ability. As deep BLSTMs are created by simply stacking multiple BLSTMs, they are very easy to construct.

In real life, there is always uncertainty, which can be described in terms of probabilities. Unlike conventional networks, which directly outpus real labels, MDN proposed by Bishop [23] can output the weighted sum of multiple probability distributions. Graves [24] showed detailed MDN calculation. Additionally, he used RNN and MDN for handwriting synthesis. The results showed that ordinary people could hardly distinguish personal handwriting from the artificially generated version. Zen et al. [25] used deep MDN to solve the problem of generating multiple outputs in speech synthesis to achieve more natural sound. The paper published in ICLR 2017 by Bazzani [26] made further significant contributions. They modeled visual attention with a mixture of Gaussians at each frame. Without a priori knowledge, the model could recognize human behavior in the video, an ability that can be leveraged to improve the accuracy of baseline action classification.

Based on this previous research, we propose a new model, deep BLSTM-MDN. In the next section, we describe the principles and structure of deep BSLTM-MDN, and then introduce the training process. In the third section, we present the experimental process and compare different models in hit-or-miss classification performance. Finally, based on the proposed model, we generated the basketball trajectory. In the fourth section, we summarize the contribution and the limitations of this approach, and make suggestions for future research.

## 2 Deep BLSTM-MDN

In this section the motivation behind the BLSTM-MDN model is described. After the structure of model is elaborated, we explain how layers are stacked and the reshaped parameters that result. Finally, details about training and future directions are discussed.

## 2.1 Motivation

LSTM is a variant of RNN, which has the same type of input and output. But in contrast to RNN, LSTM has an input gate, a forget gate, and an output gate. Thus, it can control what needs to be preserved and what needs to be forgotten. This is why LSTM can retain information from a long time ago, whereas RNN cannot. BLSTM is derived from LSTM. Its main idea is that the output of each layer can process information from both forward units and backward units. A simple example is that when we want to understand the meaning of a word, the best way is to guess it according to the context.

In contrast to conventional BLSTM networks, which only give a point estimation for the target data, BLSTM-MDN outputs a subset of values based on their probability density function (PDF). The appropriate number of PDFs needs to be specified initially. Even though the weighted sum of PDFs can, in principle, represent any arbitrary distributions, too many PDFs will in fact lead to overfitting. Weights are calculated through maximum likelihood estimation.

Given the valuable features of both BLSTM and MDN, we offer the deep BLSTM and MDN model. It has the following advantages: 1) It does not require feature extraction; 2) Deep BLSTMs have strong spatiotemporal learning ability; 3) MDN can represent the true probability of the full probability distribution; 4) It can generate the basketball trajectory in three dimensions.

## 2.2 Structure

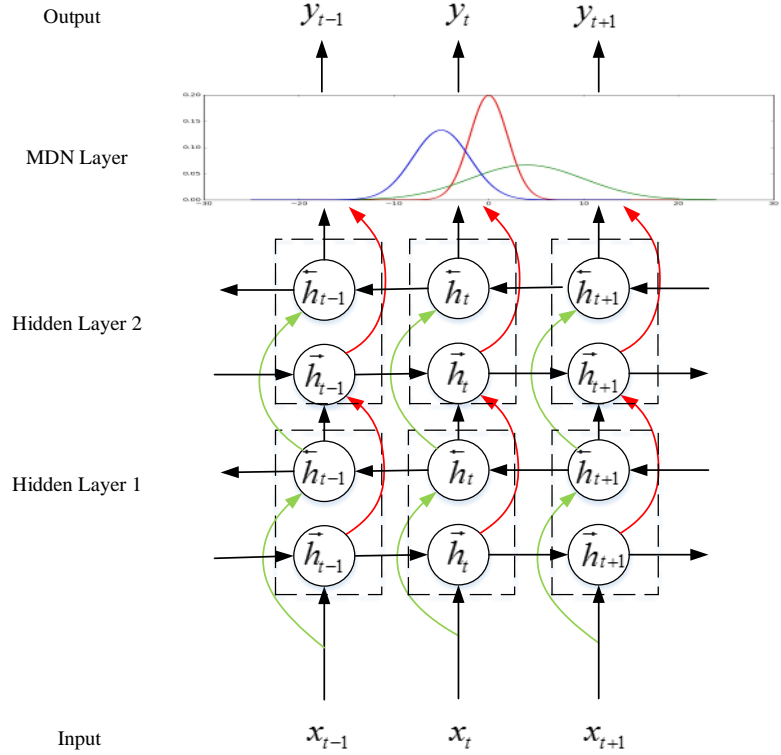

Figure 1: Unfolded structure of Deep BLSTM and MDN

In Fig. 1, the network has two hidden layers, and the output comprises three weighted MDNs. The horizontal arrows represent bidirectional flow in the temporal axis. The vertical arrows represent the one-way flow from input layer to hidden layer and from hidden layer to output layer. The red lines and green lines are forward and backward unit flows, respectively.

Let $S = \{(y^i, \mathbf{x}^i)\}_{i=1}^{N}$ represent the set of $N$ samples. For the sample $i$, the input $\mathbf{x}^i$ has four features: trajectory in three dimensions, and a time clock. But the output $y^i$ depends on different tasks. For the hit-and-miss classification task, $y^i$ has a binary value of hit and miss. For the generation task, $y^i$ is the evaluation of the next point $\mathbf{x}^{i+1}$.

LSTM adds three gates to avoid gradient vanishing and to improve memory over long periods of time. The input gate ($i_t$), forget gate ($f_t$), output gate ($o_t$), internal memory ($c_t$), and LSTM unit output ($h_t$) at time step $t$ are computed as follows:

$$\begin{aligned}
f_t &= \sigma(W_{xf} x_t + W_{hf} h_{t-1} + b_f) \\
i_t &= \sigma(W_{xi} x_t + W_{hi} h_{t-1} + b_i) \\
o_t &= \sigma(W_{xo} x_t + W_{ho} h_{t-1} + b_o) \\
c_t &= f_t \odot c_{t-1} + i_t \odot tanh(W_{hc} h_{t-1} + W_{xc} x_t + b_c) \\
h_t &= o_t \odot tanh(c_t)
\end{aligned} \quad (1)$$

where $\sigma(\cdot)$ is the sigmoid activation function, and $\odot$ is elementwise multiplication.

For clarity, $LSTM(\cdot)$ is used to represent all LSTM's functions from eqn 1. Therefore, a single BLSTM layer can be concatenated with a forward sequence and a backward sequence.

$$\begin{aligned}
\vec{h}_t &= LSTM(x_t, \vec{h}_{t-1}) \\
\overleftarrow{h}_t &= LSTM(x_t, \overleftarrow{h}_{t+1}) \\
y_t &= g(W_{\vec{h}y} \vec{h}_t + W_{\overleftarrow{h}y} \overleftarrow{h}_t + b_y)
\end{aligned} \quad (2)$$

where $g(\cdot)$ represents the activation function, $W_{mn}$ represents the weight from $m$ to $n$, and $b_n$ represents bias at layer $n$. Additionally, we choose rectified linear unit (ReLU) [28] as the activation function.

For the output layer, the Gaussian function is chosen as PDF, which can be defined as

$$P(y_t | N_t) = \sum_{c=1}^{C} \omega_t^c N(y_t | \mu_t^c, \sigma_t^c, \rho_t^c), \quad (3)$$

where $y_t$ is the ground truth, $C$ is the number of PDF, $\omega_t^c$ is the weight of the $c^{th}$ PDF, and $N(\cdot)$ is the Gaussian function.

Because the probability distribution should be valid, the parameters in $N(\cdot)$ are normalized as follows:

$$\begin{aligned}
\mu_t^c &= \tilde{\mu}_t^c \\
\omega_t^c &= \frac{\exp(\tilde{\omega}_t^c)}{\sum_{i=1}^{C} \exp(\tilde{\omega}_i^c)}, \\
\sigma_t^c &= \exp(\tilde{\sigma}_t^c) \\
\rho_t^c &= \tanh(\tilde{\rho}_t^c)
\end{aligned} \quad (4)$$

where $\tilde{\mu}_t^c$, $\tilde{\sigma}_t^c$, $\tilde{\omega}_t^c$ and $\tilde{\rho}_t^c$ are mean value of output, variance of output, weight of the PDF, and the correlation of the $c^{th}$ Gaussian component, respectively.

### 2.3 Training

To make the output as close as possible to the ground truth, we need to maximize the probability likelihood:

$$L(\mathbf{x}) = \sum_{t=1}^{T} -\log\left(\sum_{c=1}^{C} \omega_t^c N(y_t \mid \mu_t^c, \sigma_t^c, \rho_t^c)\right), \qquad (5)$$

where $T$ is the sequence length. To calculate the loss function, we choose the Adam optimizer function. The Adam optimizer can improve the traditional gradient by using momentum (the moving average of the parameters). To simplify the calculation, we assumed that the trajectory in the z-axis was uncorrelated with that in the x- and y-axes.

## 3 Experiment

The computer used to test the model had an i7 CPU with 8 GB RAM and an NVIDIA GTX 960 m GPU, which has 640 CUDA cores and 4 GB RAM. Both GPU and CPU were used, depending on the size of the neural network, which sometimes exceeded the available amount of memory on the graphics card during training.

### 3.1 Data Pre-processing and Tricks

SportVu is an optical tracking system that can record the spatial position of the ball and players on the court 25 times a second during a game. We used the NBA 2015–2016 season, a total of 631 games with 20780 three-point shots, as data for the experiment. For the shots, most sequence lengths ranged from 30 to 70 points. To deal with the unequal lengths, the model disregards the shots that are less than 12 points in length, and it cuts off the others at 12 points. This reveals that the eliminated shots account for 1.04% only of all shots, which has a negligible effect on the final results. 12 points is a reasonable number, as a few points relative to the basket determine whether each shot will be a hit or a miss; additionally, it makes efficient use of the computing resources.

We define the tensor $[X, Y, Z, T]$ as input, where $X$ refers to the length of the court, $Y$ is the width of the court, $Z$ is the height of the ball, and T is the game clock at the point of measurement. For the first (hit or miss) task, the output is a hit or miss on one shot. For the second (trajectory generation) task, the output is the next location of the ball over time in three dimensions. In the experiment, we set the basket as the center point, and transformed each point in sequence to a relative value. The dataset was divided into a training set and a test set based on Pareto principle.

When the model has been trained for several epochs, it begins to overfit. Therefore, we used an early stop to overcome the problem. This will stop the process if the present loss is lower than 90% of the mean value of the last 10 losses. In contrast to the trial-and-error method, we used grid search [29] to decrease the range of hyper-parameters, and then used the Python Hyperopt library [30] to find the best values. The Hyperopt library provides a parallel solution for model selection and parameter optimization in Python. Hyperopt is like a black box, where users only need to input an evaluation function and parameter space, and from these, they can obtain the best value within the space. In this paper, we chose the Tree of Parzen Estimators (TPE) algorithm as an optimization algorithm.

**3.2 Hit or Miss**

In this section, we describe how we tested several models for the hit or miss classification task and analyzed the result. The hit rate for the whole dataset was 35.7%, which means that the two classes (hit and miss) were disproportionally represented. So the area under the curve (AUC) was a more appropriate tool for evaluating the different models' performance.

For GLM and GBM, extra information is required in the form of additional variables based on the physics of ball trajectories, such as the angle of the ball with respect to the rim, and the distance to the rim. We believe that the additional information can improve the classification accuracy of the models.

For the other models, we built two hidden layers. Specifically, we set three Gaussian functions as MDN for LSTM-MDN and BLSTM-MDN. The epoch is 300, and it loops 50 times, with Hyperopt set for the best AUC.

Table 1: AUC at different distances to the basket for each model

| Distance to basket | Previous models [14] | | | Our models | | | |
|---|---|---|---|---|---|---|---|
| | GLM | GBM | RNN | CNN | LSTM-MDN | BLSTM | BLSTM-MDN |
| 2 feet | 0.875 | **0.942** | 0.930 | 0.800 | 0.923 | 0.926 | 0.933 |
| 3 feet | 0.807 | 0.902 | 0.913 | 0.796 | 0.918 | 0.924 | **0.925** |
| 4 feet | 0.721 | 0.848 | 0.906 | 0.790 | 0.915 | 0.918 | **0.922** |
| 5 feet | 0.659 | 0.796 | 0.880 | 0.763 | 0.887 | 0.890 | **0.910** |
| 6 feet | 0.604 | 0.746 | 0.873 | 0.751 | 0.877 | 0.879 | **0.903** |
| 7 feet | 0.583 | 0.742 | 0.841 | 0.748 | 0.848 | 0.863 | **0.882** |
| 8 feet | 0.558 | 0.719 | 0.843 | 0.740 | 0.846 | 0.851 | **0.869** |

The best AUC is marked in bold (table 1). Generally, the closer to the basket, the better is the AUC. For the same distance, deep learning models perform better than conventional models (GLM and GBM). However, convolutional neural network (CNN) performs no better than conventional models for this exercise, even though it performs well in many other fields. BLSTM-MDN gets the best AUC at 3–8 feet (table 1).

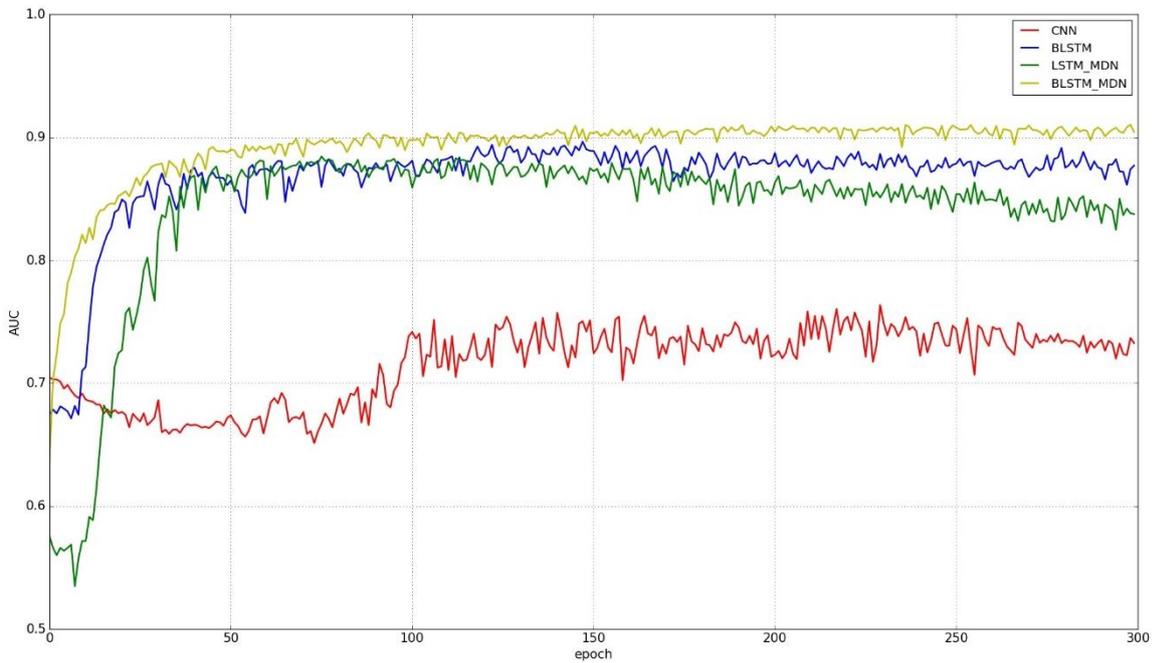

Figure 2: AUC for models at 5 feet from the basket

Table 2: Performance of deep learning models at 5 feet

| Model | AUC | Best epoch | Variable number | Spend time (average) per second |
|---|---|---|---|---|
| CNN | 0.763 | 229 | 2532 | 300 |
| LSTM-MDN | 0.887 | 75 | 52378 | 2357 |
| BLSTM | 0.890 | 147 | 34434 | 1642 |
| BLSTM-MDN | **0.910** | 298 | 35994 | 1859 |

To compare the performance of the models in more detail, we selected the distance of 5 feet to analyze the AUC for each epoch. In Fig. 2, CNN, BLSTM, LSTM-MDN, and BLSTM-MDN are drawn with red, blue, green, and yellow lines, respectively. Initially, BLSTM, LSTM-MDN, and BLSTM-MDN had the same upward trend, the AUCs increased sharply, and they started to oscillate from epoch 40 at close range. CNN had the same initial value as the others, but the AUC increased much more slowly. The best AUC was around epoch 100. Overall, BLSTM-MDN performed both well in the convergence rate and final AUC.

Table 2 shows the AUC, epoch, variable number, and time spent for each model. It should be noted that each model was re-run 50 times with Hyperopt, and the table shows the best AUC. The four models all have two hidden layers, and each layer has 64 units. CNN has the simplest structure, with only 2532 variables, and only 300 second-run cost per run with Hyperopt. The BLSTM model splits hidden units into forward LSTMs and backward LSTMs on average. In other words, each LSTM has 32 units. Therefore, the number of variables in BLSTM-MDN has fewer variables than LSTM-MDN. This means that running BLSTM-MDN costs less than running LSTM-MDN does (1859 vs. 2357 variables).

### 3.3 Trajectory Generation

Using BLSTM-DMN, we generated the trajectory of a flying basketball. The model has the same hyper-parameters as were used for section 3.2. Thanks to MDN, it can obtain the distribution probability of each point. Therefore, it is easy to know the next point using a roulette algorithm.

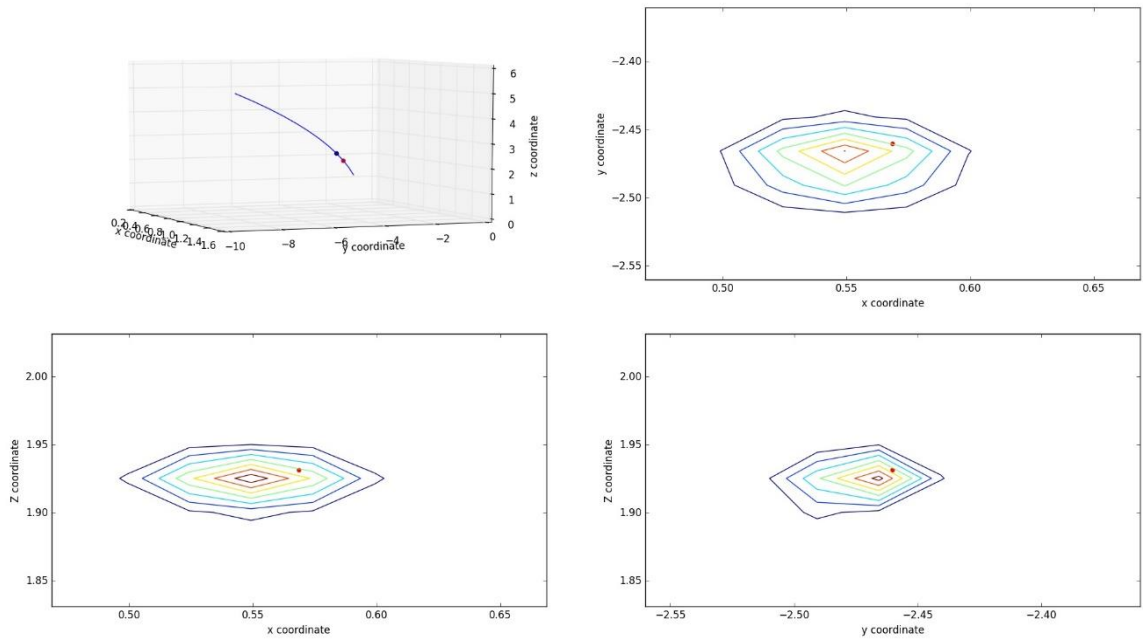

Figure 3: Distribution probability of the 10$^{th}$ point given the 9$^{th}$ point

The top left subplot in Fig 3 shows a real 3D trajectory, which is composed of 12 discrete points. The red point and blue point denote the 9$^{th}$ and 10$^{th}$ points respectively. The other subplots are 10$^{th}$ point in the horizontal plane, vertical plane and width plane respectively. The circles around mean the probabilistic contour lines of the 10$^{th}$ point. And probability increase along with the color ranging from blue to red. It is obviously that the real position (red point) is almost locates in the center of each circle, which means BLSTM-MDN gets very high accuracy prediction.

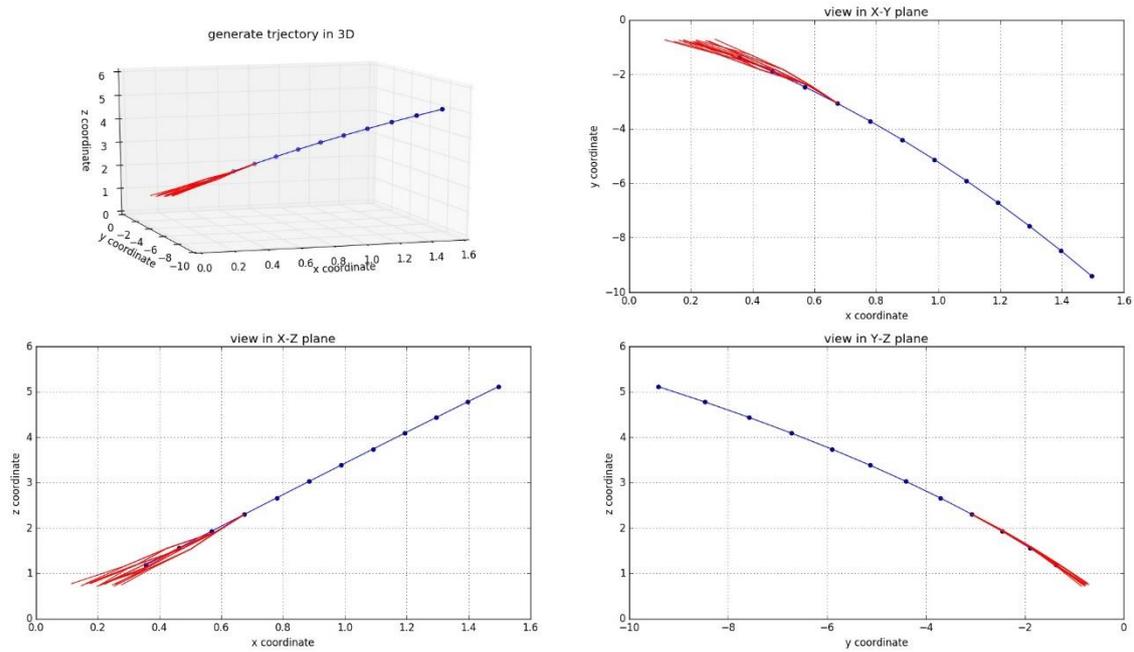

Figure 4: Generated trajectories with given time stamps

In Fig. 4, the blue lines are real trajectories, and the red ones are generated. Again, we picked the $9^{th}$ point as the time stamp. Then, for the subsequent positions (the $10^{th}$, $11^{th}$, and $12^{th}$), the model generated two predicted points. In sum, the model had eight ($2^3$) trajectories. It can be seen that the generated trajectories are very similar to the real one.

## 4 Conclusion

We introduced a BLSTM-MDN model for three-point shot prediction. During model training, the Python library Hyperopt was applied for hyper-parameter self-optimization. Compared to the previous models, such as the conventional CNN and BLSTM models, the proposed one performed better in both convergence rate and prediction accuracy.

Even though the basketball trajectory prediction is a toy problem, we suggest that BLSTM-MDN can produce correct results in many other areas, such as UAV route planning, human activity recognition, and stock market prediction. There are also many factors that need to be further considered. For example, the model should be able to take into account player cooperation and defense when predicting NBA player positions. Time series prediction is a complicated but meaningful research topic. We aim to improve our model by taking into account more features in the future.